\documentclass[runningheads]{llncs}
\usepackage[T1]{fontenc}
\usepackage{graphicx}
\usepackage{tabularx}
\usepackage{multirow}
\usepackage{booktabs}
\usepackage{hyperref}
\usepackage[table]{xcolor}
\usepackage{amsmath}
\usepackage{array}
\newcolumntype{R}[1]{>{\raggedleft\arraybackslash}p{#1}}
\usepackage{enumitem}
\begin{document}
\title{Scalable and Explainable Learner-Video Interaction Prediction using Multimodal Large Language Models}
\titlerunning{Video Interaction Prediction with Multimodal Large Language Models}
\author{Dominik Glandorf\inst{1}\orcidID{0009-0000-0038-2721} \and Fares Fawzi\inst{1}\orcidID{0000-0002-5331-579X} \and
Tanja Käser\inst{1}\orcidID{0000-0003-0672-0415}}

\authorrunning{D. Glandorf et al.}

\institute{EPFL, Switzerland\\
\email{\{dominik.glandorf,fares.fawzi,tanja.kaeser\}@epfl.ch}\\
}

\maketitle              %
\begin{abstract}
Learners’ use of video controls in educational videos provides implicit signals of cognitive processing and instructional design quality, yet the lack of scalable and explainable predictive models limits instructors’ ability to anticipate such behavior before deployment. We propose a scalable, interpretable pipeline for predicting population-level watching, pausing, skipping, and rewinding behavior as proxies for cognitive load from video content alone. Our approach leverages multimodal large language models (MLLMs) to compute embeddings of short video segments and trains a neural classifier to identify temporally fine-grained interaction peaks. Drawing from multimedia learning theory on instructional design for optimal cognitive load, we code features of the video segments using GPT-5 and employ them as a basis for interpreting model predictions via concept activation vectors. We evaluate our pipeline on 77 million video control events from 66 online courses. Our findings demonstrate that classifiers based on MLLM embeddings reliably predict interaction peaks, generalize to unseen academic fields, and encode interpretable, theory-relevant instructional concepts. Overall, our results show the feasibility of cost-efficient, interpretable pre-screening of educational video design and open new opportunities to empirically examine multimedia learning theory at scale.

\keywords{Video-based learning \and Multimodal large language models \and Explainable AI \and Predictive modeling}
\end{abstract}
\section{Introduction}
The rise of digital and asynchronous learning has made educational videos a ubiquitous teaching medium \cite{sablic_video-based_2021}. Millions of learners engage with video-based online learning environments such as massive open online courses (MOOCs), university-run flipped classrooms, and even general-purpose video platforms \cite{papadakis_moocs_2023,navarrete_closer_2025}. Unlike live lectures, instructors receive no direct feedback on which moments challenge learners or impose unnecessary cognitive load. Tools that identify such moments can therefore support instructional video design and iterative improvement \cite{chavan_tcherly_2022}.

The Cognitive Theory of Multimedia Learning (CTML) \cite{mayer_principles_2014} is a common framework for optimizing instructional video design. It highlights cognitive load as a central variable and specifies \textit{multimedia features}, such as visual complexity, redundancy, human presence, information presentation modalities, and segmentation, that affect cognitive processing demands, and it formulates principles for effective multimedia instruction.

Empirical studies have related variation in these features \textit{within} instructional videos to learner interaction behavior as a proxy for cognitive load. They indicate that learners pause and rewind at moments of increased processing difficulty or conceptual and visual boundaries \cite{merkt_pushing_2022,kim_understanding_2014} and that higher textual and graphical complexity is associated with increased pausing, rewinding, and dropout \cite{gritz_formulas_2025,atapattu_impact_2018}.
These analyses typically rely on \textit{manual annotation} of multimedia features, which is labor-intensive and difficult to scale, especially at the temporal resolution of individual video moments. Moreover, they are primarily descriptive: while they identify associations between CTML features and learner behavior, they do not provide predictive models that can estimate how learners will interact with new instructional videos prior to deployment.

In parallel, a large body of work has focused on predicting learning outcomes and engagement from learners’ video interaction data. Most prior studies rely on post-hoc outcome prediction (e.g.,~\cite{lalle2020data}), with comparatively few predicting intermediate outcomes during a course or video interactions~\cite{mbo20early,swamy_meta_2022}. The features used in these models range from aggregated video control statistics (e.g., number of clicks or average pause duration; e.g.,~\cite{lalle2020data}) to more fine-grained interaction patterns (e.g.,~\cite{akpinar_2020,li_mooc_2015}), but are generally content-agnostic. Work that explicitly incorporates content typically relies on high-level course metadata~\cite{swamy_meta_2022} or treats videos as monolithic units rather than sequences of moments~\cite{asadi_ripple_2023,zhang_video_2022,stohr_videos_2019,thornton_criteria_2017,shoufan_estimating_2019}. Even studies that model positions or sequences of activities within videos do not explicitly link theory-driven multimedia features to learner behavior~\cite{brinton_mining_2016,sinha_your_2014,lee_playback-centric_2022,li_mooc_2015}.

As a result, predictive models of learner-video interactions that operate at moment-level granularity, use raw video content, and are grounded in multimedia learning theory remain largely unexplored. Most institutions do not systematically map large-scale video-interaction data to CTML features, nor do they reach sufficient sample sizes to draw timely conclusions about instructional quality \cite{lee_playback-centric_2022}. This gap highlights the need for a scalable, explainable approach to address the cold-start problem in behavioral feedback and support data-driven video design.

In this paper, we propose a scalable and explainable pipeline for predicting design-related peaks in learner-video interactions from video content alone. Our approach uses multimodal large language models (MLLMs) to compute video embeddings and trains a neural classifier to predict moment-to-moment watching, pausing, rewinding, and skipping behavior based on video content alone. To interpret these predictions, we draw on multimedia learning theory by coding CTML features with GPT-5 and applying Testing with Concept Activation Vectors (TCAV) to assess how strongly these features are encoded in the embeddings and influence model predictions. Accordingly, we address three research questions: \textbf{(RQ1)} to what extent moment-to-moment learner-video interactions can be predicted from video embeddings; \textbf{(RQ2)} whether these predictions can be explained using AI-coded CTML features; and \textbf{(RQ3)} how sensitive embedding-based predictions are to such features.

We evaluate our approach on a multi-year dataset of over 77 million video interactions across 1{,}641 videos from 66 courses. Our results show that embedding-based models achieve strong predictive performance, generalize to unseen academic fields, and encode interpretable, theory-relevant CTML features. While CTML features alone are less predictive than embeddings, the TCAV analysis demonstrates that embeddings meaningfully capture multimedia learning concepts at low computational cost.
In summary, leveraging recent advances in video understanding, our approach enables scalable, CTML-grounded prediction of moment-to-moment learner-video interactions, enabling cost-efficient, interpretable pre-screening of instructional video design and large-scale empirical investigation of multimedia learning theory. We publicly release the models and coding scheme, including prompts, along with the code and supplementary material\footnote{\scriptsize\url{https://github.com/epfl-ml4ed/video-interaction-prediction-AIED26}}.

\section{Method}
\vspace{-1mm}
\begin{figure}[t]
  \centering
  \includegraphics[width=\linewidth, trim=1.5cm 12.25cm 4cm 3cm, clip]{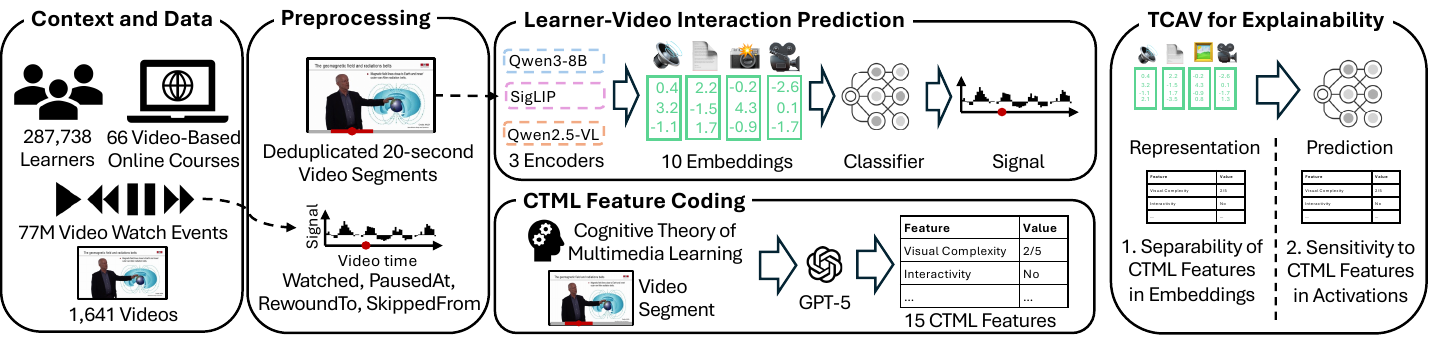}
  \caption{Pipeline for predicting learners’ interactions with online learning videos and explaining predictions from multimedia learning theory.}
  \label{fig:pipeline}
  \vspace{-2mm}
\end{figure}

We propose a multi-step pipeline for predicting aggregated learners’ video interaction behavior (pausing, rewinding, watching, and skipping) on unseen instructional videos in an interpretable manner (see Fig.~\ref{fig:pipeline}). First, we aggregate learner-video interactions into behavioral signals over the video timeline. Next, we employ MLLMs to produce embeddings of video content and train a neural network to predict these behavioral signals from the content alone. Finally, to explain the model’s predictions, we use GPT-5 to encode video content with CTML-based features and apply TCAV for interpretation.

\vspace{-2mm}
\subsection{Context and Data}
\label{sec:data}
\vspace{-1mm}
We evaluated our approach on a large-scale dataset of learner clickstreams collected between 2013 and 2025 from a video-based learning platform supporting MOOCs and flipped classroom courses. The platform offers courses from a technical university to both worldwide online learners and enrolled on-campus students. The courses span Science, Technology, Engineering, and Mathematics (STEM) and related fields, and many were adapted from traditional lectures and redesigned for web-based instruction. The institutional ethics review board approved the study (Approval No. 000612).

We included courses with at least one hour of video material in which at least 200 active learners interacted, yielding 174 runs of 66 distinct courses (67\% English, 33\% local language) from 11 fields. The thresholds were chosen to balance sufficient signal quality of interaction behavior and the inclusion of less popular courses. Table~\ref{tab:data} presents detailed course statistics. Our dataset comprised 77.3 million video interactions from 357,265 course participants (287,738 unique). Interaction types were distributed as follows: plays (44.2\%), pauses (27.7\%), seeks (23.7\%), and play rate changes (1.1\%). Among seek events, 43.9\% were rewinds and 56.1\% were skips. The data comprised 1,641 unique videos, out of which 1,322 appeared in at least two course runs. The total length of the videos was 347.2 hours.

\begin{table*}[t]
  \caption{Statistics of online learning data in our study. "Others" include Architecture, Biology, Computer Science, Education, Environmental Science, and Microengineering.}
  \label{tab:data}
  \center
  \vspace{-5mm}
\resizebox{0.98\textwidth}{!}{%
\begin{tabular}{lrrrrrr}
\toprule
\textbf{Field} & \textbf{\#Courses} & \textbf{\#Course runs} & \textbf{\#Learners} & \textbf{Avg. \#Videos} & \textbf{Avg. Length [min]} \\
\midrule
Mathematics & 18 & 42 & 49,311 & 62.7 & 12.1  \\
Neuroscience & 15 & 28 & 78,067 & 43.7 & 13.5  \\
Electrical Engineering & 10 & 27 & 43,841 & 41.9 & 17.9  \\
Physics & 10 & 18 & 60,924 & 37.8 & 17.0  \\
Management & 9 & 15 & 19,451 & 44.3 & 10.9  \\
Others & 23 & 44 & 105,671 & 41.6 & 12.0  \\
\bottomrule
\end{tabular}
}
\vspace{-2mm}
\end{table*}

\vspace{-2mm}
\subsection{Preprocessing}
\label{sec:signal_processing}
\vspace{-1mm}
We preprocessed both the videos and video interaction events.

\vspace{1pt} \noindent \textbf{Videos}
For each video $v$, we extracted its duration $D_v$. We deduplicated videos using $D_v$ and the hash of a resized thumbnail at a fixed time point. The platform operator provided us with OCR-extracted slide texts and automatically generated transcripts, both machine-translated into English when necessary.

\vspace{1pt} \noindent \textbf{Learner Interactions} We aggregated learner interactions from logged video events (play, pause, rewind, skip) within each video to construct ground-truth prediction targets, yielding four continuous population-level signals ${\text{Signal}}_v(t)\in [0,1]$ for each rounded video second $t$: ${\text{Watched}}_v(t)$ (Total number of views of $t$), ${\text{PausedAt}}_v(t)$ (Number of pauses at $t$), ${\text{RewoundTo}}_v(t)$ (Number of backward seeks to $t$), and ${\text{SkippedFrom}}_v(t)$ (Number of forward seeks from $t$). These signals capture distinct forms of content interaction at time $t$, assuming that rewinding relates to its destination and skipping to its origin. We omitted playrate changes due to their scarcity. To isolate content-specific effects, we adapted preprocessing from prior work~\cite{gritz_formulas_2025,kim_understanding_2014}:

1) We removed the first and last 30 seconds of each video to exclude content-independent patterns (see \href{https://github.com/epfl-ml4ed/video-interaction-prediction-AIED26/blob/main/supplementary_material.pdf}{supplement}); 2) We normalized by the number of active learners at $t$ to control for in-video dropouts; 3) we smoothed signals using a five-second moving average to account for delayed reactions; and 4) we linearly detrended the signal to remove global temporal trends (e.g., position-in-video effects) that are unrelated to video content. Finally, 5) we defined the signals in terms of percentile ranks: ${\text{Signal}}_v(t)=\frac{\text{rank}_{t\in\{30,...,D_v-30\}}(\text{Events}_v(t))}{D_v-60}$. We normalized signal magnitudes in this way to identify relative peaks across the video timeline, corresponding to time points where comparatively more learners exhibited a given behavior (e.g., pausing). We adopted rank-based normalization because the event signals were not normally distributed (see \href{https://github.com/epfl-ml4ed/video-interaction-prediction-AIED26/blob/main/supplementary_material.pdf}{supplement}) and because it standardizes signal magnitudes across videos.

\vspace{-3mm}
\subsection{Learner-Video Interaction Prediction}
\label{sec:watching-pred}
\vspace{-1mm}

\begin{figure}[t]
  \centering
  \includegraphics[width=\linewidth, trim=4.9cm 12.7cm 3.85cm 2.75cm, clip]{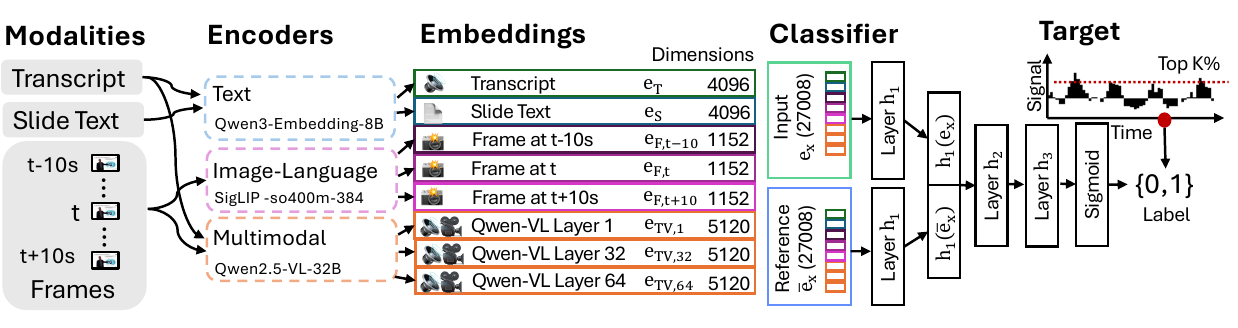}
  \caption{Three modalities of video segments around $t$ are encoded by pre-trained transformers. A neural classifier predicts if ${\text{Signal}}_v(t)$ is among the top K\% at timepoint $t$.}
  \label{fig:architecture}
\end{figure}

Our goal is to predict learner-video interactions from video content. Fig.~\ref{fig:architecture} visualizes the classification pipeline. We define the binary target as $y_v(t)=1$ if ${\text{Signal}}_v(t)\ge 1-K\%$, meaning the signal magnitude at $t$ is among the top K\% per video. The task is then to predict $y_{v}(t)$ from a 20-second segment $x_{v,[t-10,t+10]}$, to include sufficient video context that contributes to ${\text{Signal}}_v(t)$ but discard irrelevant information. To assess sensitivity to the segment length, we also include the main results using a 10-second segment in the \href{https://github.com/epfl-ml4ed/video-interaction-prediction-AIED26/blob/main/supplementary_material.pdf}{supplement}. Although a learner may not have watched the video beyond $t$, a symmetric segment around $t$ may still help the model understand the context. In the following, we omit the subscript of $x$ for simplicity.

\vspace{1pt} \noindent \textbf{Embeddings from MLLMs.}
\label{sec:embeddings} MLLMs show state-of-the-art video understanding capabilities; their encodings likely contain information relevant to identifying critical video moments. Therefore, we precomputed embeddings of video segments encoded by an MLLM, and as baselines, a text-only LLM and an image-language model. Precisely, we computed contextualized dense vector representations of $x$, comprising three modalities: the transcript segment $T_{[t-10,t+10]}$, slide text $S_t$, and a temporal window of video frames around $t$. Let $\mathcal{F}_t = \bigl(F_{t-10}, \dots, F_{t+10}\bigr)$ denote the tuple of 21 discrete video frames sampled at 1 frame per second in a window centered at $t$.

We encoded $T_{[t-10,t+10]}$ and $S_{t}$ using a state-of-the-art text embedding model (\texttt{Qwen3-Embedding-8B}~\cite{yang_qwen3_2025}), yielding embeddings $e_{T}$ and 
$e_S$, respectively. For visual content, we applied two complementary encodings to the same frame window $\mathcal{F}_t$. First, each frame $F_i \in \mathcal{F}_t$ was independently encoded using a transformer pre-trained for image-text alignment (\texttt{SigLIP-so400m-patch14-384}~\cite{zhai_sigmoid_2023}), yielding per-frame embeddings $e_{F_i}$, where $e_{F,t}$ denotes the embedding of the frame at time $t$. We then formed a sparse visual embedding $e_{F,(t-10,t,t+10)}$ by concatenating the embeddings of frames $\bigl(F_{t-10}, F_t, F_{t+10}\bigr)$. Second, we encoded the full ordered frame sequence $(F_{i})_{i=t-10}^{t+10}$ using a large vision-language model (\texttt{Qwen2.5-VL-32B-Instruct}~\cite{bai_qwen25-vl_2025}) without an explicit instruction prompt. We extracted hidden representations from layers $l \in \{1, 32, 64\}$ to capture multiple encoding depths and applied token-wise mean pooling. 
Since this model does not process audio, we computed two variants of this embedding: one with the transcript prepended to the frame sequence, denoted $e_{TV,l}$, and one without the transcript, denoted $e_{V,l}$. We additionally computed $e_{TV,l,7B}$ using the 7B variant of the encoder. 
The entire computation required $\sim$100 GPU hours on a single NVIDIA A100-80GB GPU.

\vspace{1pt} \noindent \textbf{Classifier.}
\label{sec:classifier}
We trained a lightweight classification head $H$ to predict $y(t)$. First, a linear dense layer $h_1$ was applied to the segment embedding $e_x$. In the default setting, $e_{x}$ was formed by concatenating $e_{T}$, $e_{S}$, $e_{F,(t-10,t,t+10)}$, and $e_{TV,\{1,32,64\}}$. 
To provide a video-level description of content, which may be useful to rank moments against each other, we precomputed the mean embedding $\bar e_{{x\in v}}$ over all segments in video $v$ (``Reference") and applied the same layer $h_1$ (``Shared Weights"). A second dense layer $h_2$ received the concatenation $\bigl(h_1(e_{x}),h_1(\bar e_{{x\in v}})\bigr)$. $h_3$ projected the output of $h_2$ to a single logit, which was passed through a sigmoid activation to obtain the predicted class probability. We used a binary cross-entropy loss and ReLU activation functions. The hidden dimension of $h_1$ and $h_2$ was fixed to 256 after observing sufficient capacity in a preliminary experiment, yielding a total model size of 7M parameters.

\begin{table*}[t]
  \caption{Rubric for coding CTML features of the video lecture content.}
  \label{tab:coding_scheme}
{\fontsize{7}{8}\selectfont
\rowcolors{3}{white}{gray!7}
\begin{tabularx}{\textwidth}{p{1.5cm}p{1.7cm}Xp{0.7cm}p{0.8cm}}
\toprule
\textbf{Modality} & \textbf{Feature} & \textbf{Description} & \textbf{Scale} & \textbf{Lit.} \\
\midrule
$F_{t}$ & Formula & Math notation beyond single symbols is visible. & \{0,1\} & \cite{gritz_formulas_2025}   \\
$F_{t}$ & Instructor & Instructor's head is visible in the video. & \{0,1\} & \cite{fiorella_instructor_2019}   \\

$F_{t}$ & Screen & The user interface of a computer screen is embedded inside the image (e.g., a code editor). & \{0,1\} & \cite{gritz_formulas_2025}   \\
$F_{t}$ & Structured Information Visualization & A diagram, graph, schematic drawing, or table is on the slide (no formulas, no GUI). & \{0,1\} & \cite{gritz_formulas_2025}   \\
$F_{t}$ & Text Object & A printed or handwritten sentence, derivation, or bullet point is visible (no labels, titles, footers, text in screenshots or code editors, references). & \{0,1\} & \cite{gritz_formulas_2025}   \\
$F_{t}$ & Visual Complexity & The amount, complexity, and diversity of textual and graphical content objects visible. Ignore instructors and typical slide elements like logos or titles as elements. & 1-5 & \cite{gritz_formulas_2025}   \\
$\bigl(F_{{i}}\bigr)_{i=t-10}^{t+10}$ & Annotating & In the frame progression, something new is being written by hand on the slide or typed letter by letter inside an editor (no pens moving without writing). & \{0,1\} & \cite{fiorella_instructor_2019}   \\
$\bigl(F_{{i}}\bigr)_{i=t-10}^{t+10}$ & Animation / Video & Embedded videos or built-in animations (e.g., objects, figures, or video footage). Frames showing only lecturer movement, keyboards, hands, pointers, showing or typing or writing new text, or slide transitions do not meet this requirement. & \{0,1\} & \cite{kuhlmann_students_2024}   \\
$\bigl(F_{{i}}\bigr)_{i=t-10}^{t+10}$ & Photo & The second frame shows or includes a static real-world photograph, e.g., nature, people, objects, scenery (no instructor, hands, keyboard, or pointer, scans of handwritings). & \{0,1\} & \cite{gritz_formulas_2025}   \\
$\bigl(F_{{i}}\bigr)_{i=t-10}^{t+10}$ & Showing & The lecturer's hand or pen appears on the slide (no writing or speaking hand gestures). & \{0,1\} & \cite{chorianopoulos_taxonomy_2018} \\
$\bigl(F_{{i}}\bigr)_{i=t-10}^{t+10}$ & Visual Breakpoint & Clear slide transitions or cuts in videos and animations. No additions to the current slide, focus on disappearing content. & \{0,1\} & \cite{kim_understanding_2014,merkt_pushing_2022} \\
$T_{[t-10,t+10]}$ & Signaling & (Subtle) hints of the importance of information (``main", ``important", ``interesting", ``key", ``noteworthy", etc.). & \{0,1\} &  \cite{mayer_principles_2014}   \\
$T_{[t-10,t+10]}$ & Interactivity & Questions or prompts to the audience and other suggestions to active learning (such as reflections, lookups, or exercises). & \{0,1\} & \cite{freeman_active_2014}  \\
$T_{[t-10,t+10]}$, $F_{t}$ & Semantic Breakpoint & The video could be clearly cut at one point where the speaker starts a new point, an example, a summary, an enumeration, a side note. & \{0,1\} & \cite{merkt_pushing_2022,fiorella_instructor_2019}  \\
$T_{[t-10,t+10]}$, $\bigl(F_{{i}}\bigr)_{i=t-10}^{t+10}$ & Redundancy & Correspondence between the spoken and visible slide content. & 1-5 & \cite{mayer_principles_2014}  \\
\bottomrule
\end{tabularx}
}
 \vspace{-2mm}
\end{table*}

 \vspace{-2mm}
\subsection{CTML Feature Coding}
 \vspace{-1mm}
To explain the prediction of learner-video interactions, we required a set of well-defined CTML features applicable to this medium and potentially variable \textit{within} a video. Due to the lack of an established rubric, we defined a tailored coding scheme for CTML features of video moments, drawing from the CTML and empirical evidence. The features of our rubric are described in Table~\ref{tab:coding_scheme} with corresponding links to the literature. They comprise static visual features, such as instructor visibility or textual elements, as well as dynamic features, such as gestures, and abstract features, such as redundancy between audio and video. Each feature is linked with the modalities defined in Section~\ref{sec:embeddings}. We decided for \textit{Signaling} to be audio-only, although relevance markers could also be visual.

\vspace{1pt} \noindent \textbf{Human inter-rater agreement.} We first validated which features can be reliably coded by humans. Two human coders among the authors independently coded each feature on 24 sampled video moments, from all fields, languages, and top and non-top ranks in the interaction signals (720 ratings in total, details in \href{https://github.com/epfl-ml4ed/video-interaction-prediction-AIED26/blob/main/supplementary_material.pdf}{supplement}). The coders were presented with the feature description and the modality of $x$ indicated in Table~\ref{tab:coding_scheme}. The task was to rate the central five seconds of the video moment, as this timeframe contributes to ${\text{Signal}}_v(t)$. Cohen's $\kappa$~\cite{cohen_coefficient_1960} served as the metric of inter-rater agreement, in the quadratically weighted version for non-binary features. For ten out of 15 features, the human coders showed almost perfect agreement ($\kappa\ge0.8$); they reached substantial agreement on \textit{Redundancy} ($\kappa\ge0.6$). The features \textit{Signaling}, \textit{Interactivity}, and \textit{Semantic Breakpoint} were more challenging to agree on ($\kappa\ge0.4$). The first two features varied in the subtlety of signaling and suggestiveness of the interactivity. The presence of a semantic breakpoint may need more context than a short excerpt from the video transcript. We provide the $\kappa$ per feature in the \href{https://github.com/epfl-ml4ed/video-interaction-prediction-AIED26/blob/main/supplementary_material.pdf}{supplement}.

\vspace{1pt} \noindent \textbf{AI inter-rater agreement.} To quantify how well this coding process can be automated and to what extent an AI understands the multimedia features, we prompted \texttt{gpt-5-2025-08-07}\footnote{\scriptsize\url{https://developers.openai.com/api/docs/models/gpt-5}} to code the same moments. We report the model's agreement with the human ratings in Section~\ref{sec:rq2}. Before this step, each human disagreement was resolved by discussion.

\vspace{-2mm}
\subsection{TCAV for Explainability}
 \vspace{-1mm}
TCAV~\cite{kim_interpretability_2018} is a method for explaining a neural network's prediction by interpreting its internal representations in terms of human-understandable concepts. 
In our context, every CTML feature naturally defines a concept $C$. TCAV allows us to assess whether a given concept is encoded in a hidden layer $l$ of the model and to quantify how the model’s classification changes when the concept is present. The CTML coding serves as the required input of examples that feature $C$.

In the first step, a linear classifier is trained to detect the presence of a concept from layer activations. As layers, we consider all input embeddings individually, as well as layer $h_1$.
The unit-normalized weights of this classifier are referred to as the concept activation vector (CAV), which can be interpreted as the direction in the activation space that represents $C$. For binary concepts, we learned CAVs using a logistic regression model, whereas for ordinal-valued concepts, we used linear regression. We tuned the L2 regularization strength for each CAV and used the same value for its random counterpart.   In the second step, a directional derivative $d^C_l(x)$ measures the local sensitivity of the model’s output logit to the concept at input $x$. As a baseline, a CAV is also trained for a random concept $C_\text{random}$. The overall sensitivity of the model to concept $C$ at layer $l$ is then summarized by the TCAV score, $TCAV^C_l = \frac{\left|\{\,x \in X^{H_+} : d^C_{l}(x) > d^{C_\text{random}}_l(x)\,\}\right|}{\left|X^{H_+}\right|}$, where $X^{H_+}$ denotes the set of inputs with positive predictions. Intuitively, $TCAV^C_l$ measures the fraction of positively predicted examples for which the model is more sensitive to concept $C$ than to a random concept.

To assess whether $TCAV^C_l$ differed from the null hypothesis of no sensitivity to concept $C$ at layer $l$, we repeated the procedure 25 times using different train-test splits for the CAVs. We then applied Welch’s $t$-test against a null value of $0.5$ to account for unequal variances, and used conservative Bonferroni corrections to control for the $15\times10$ comparisons across layer-concept combinations.
 
\vspace{-2mm}
\section{Results}
\vspace{-1mm}
We evaluated our models' performance in predicting learner-video interactions from video embeddings (RQ1), analyzed relevant information for this task in AI-coded CTML features (RQ2), and explored the relation between the predictions of embedding-based models and CTML using explainable AI (RQ3).

\vspace{1pt} \noindent \textbf{Experimental Protocol.} 
We subsampled the preprocessed signals at  5-second intervals, yielding 224,327 datapoints. We evaluated the task of distinguishing the top $K\%\in\{5,10\}$ signal moments from all remaining moments using the surrounding 20-second video context, reflecting a scale feasible for human double-checking of cognitive load (for $K=20\%$ or a 10-second context, see \href{https://github.com/epfl-ml4ed/video-interaction-prediction-AIED26/blob/main/supplementary_material.pdf}{supplement}). Performance was measured using the area under the receiver operating characteristic curve (AUC), which is insensitive to class imbalance. To assess operational usefulness, we report $\text{Lift@K\%} = \frac{\text{Precision@K\%}}{P(y(t)=1)}$ where precision is computed over the top K\% of the model scores per video and $P(y(t)=1)=K\%$ by definition. $\text{Lift@K\%}$ quantifies how much more likely a datapoint among the top K\% model scores is to be a true positive relative to a random selection.

We estimated performance using repeated random 90\%/10\% train-test splits with different seeds, ensuring that moments from the same course appeared in exactly one set to prevent leakage between courses and course runs. We computed the metrics on the full test set to reflect the true class imbalance. During training, for each $K$ and signal, we balanced the data by subsampling equal numbers of top K\% and other video moments. When computing the reference embedding $\bar e_{{x\in v}}$ on this balanced set, datapoints were reweighted by the class prevalence $K\%$ to preserve a representative reference distribution. We used 10\% of the training data for validation and early stopping, optimized with Adam (learning rate: $10^{-4}$), a batch size of 2048, and a dropout rate of $0.2$ in $h_2$. 

For a field-specific (discipline-level) evaluation, instead of using a 90-10 split, we tested on moments from courses from a held-out academic field (e.g., mathematics or engineering) and trained on all remaining courses. 

We coded CTML features with GPT-5 for 6,000 datapoints from 1,000 videos equally sampled from the top 5\% and remaining moments to ensure a sufficient number of samples in both classes. The API cost for this coding was $\sim150$USD.

\vspace{-2mm}
\subsection{RQ1: Learner-Video Interaction Prediction with MLLMs}
\vspace{-1mm}
\begin{table}[t]
\centering
\caption{Predictive performance in terms of AUC ($\pm$ std across 10 seeds) of moment-to-moment video interaction prediction from multimodal video content embeddings; Lift@K\% ($\pm$ std) is additionally reported for the final multimodal model $e_x$.}
\label{tab:modalities}
\resizebox{\textwidth}{!}{%
{
\rowcolors{3}{white}{gray!7}
\begin{tabular}{rlllllllll}

\cmidrule{1-10}
&\multicolumn{1}{r}{\textbf{Signal}$\rightarrow$}
& \multicolumn{2}{c}{\textbf{${\text{Watched}}_v(t)$}}
 & \multicolumn{2}{c}{\textbf{${\text{PausedAt}}_v(t)$}}
 & \multicolumn{2}{c}{\textbf{${\text{RewoundTo}}_v(t)$}}
 & \multicolumn{2}{c}{\textbf{${\text{SkippedFrom}}_v(t)$}} \\
\cmidrule(lr){3-4}
\cmidrule(lr){5-6}
\cmidrule(lr){7-8}
\cmidrule(lr){9-10}
&\textbf{Embeddings}$\downarrow$
& \textbf{Top 10\%} & \textbf{Top 5\%}
& \textbf{Top 10\%} & \textbf{Top 5\%}
& \textbf{Top 10\%} & \textbf{Top 5\%}
& \textbf{Top 10\%} & \textbf{Top 5\%} \\
\midrule
&$e_{T}$ & 0.58±0.02 & 0.58±0.03 & 0.62±0.02 & 0.62±0.02 & 0.59±0.02 & 0.61±0.02 & 0.58±0.02 & 0.57±0.02 \\
&$e_{S}$ & 0.50±0.01 & 0.52±0.02 & 0.53±0.01 & 0.54±0.01 & 0.56±0.03 & 0.56±0.02 & 0.57±0.03 & 0.57±0.03 \\
&$e_{F,t}$ & 0.59±0.02 & 0.59±0.03 & 0.64±0.03 & 0.68±0.03 & 0.60±0.03 & 0.62±0.04 & 0.57±0.03 & 0.57±0.02 \\
&$e_{F,(t-10,t,t+10)}$ & 0.63±0.02 & 0.63±0.04 & 0.67±0.04 & 0.70±0.04 & 0.67±0.03 & 0.69±0.06 & 0.62±0.03 & 0.62±0.05 \\
&$e_{V,\{1,32,64\}}$ & 0.65±0.03 & 0.67±0.04 & 0.71±0.03 & 0.71±0.03 & 0.71±0.02 & \textbf{0.76±0.03} & 0.62±0.04 & 0.62±0.03 \\
&$e_{T}$; $e_{S}$ & 0.56±0.01 & 0.58±0.03 & 0.62±0.02 & 0.62±0.02 & 0.60±0.02 & 0.62±0.02 & 0.59±0.04 & 0.60±0.04 \\
&$e_{T}$; $e_{F,t}$ & 0.60±0.02 & 0.60±0.02 & 0.66±0.03 & 0.68±0.03 & 0.61±0.03 & 0.62±0.04 & 0.57±0.02 & 0.59±0.03 \\
&$e_{T}$; $e_{F,(t-10,t,t+10)}$ & 0.64±0.01 & 0.68±0.03 & 0.69±0.02 & 0.71±0.05 & 0.66±0.04 & 0.69±0.07 & \textbf{0.64±0.04} & 0.61±0.03 \\
&$e_{TV,\{1,32,64\}}$ & \textbf{0.67±0.02} & \textbf{0.70±0.03} & 0.73±0.02 & 0.73±0.03 & \textbf{0.72±0.02} & 0.74±0.04 & 0.63±0.04 & 0.65±0.04 \\
&$e_x$ (all) & \textbf{0.67±0.02} & 0.69±0.03 & \textbf{0.74±0.02} & \textbf{0.74±0.02} & \textbf{0.72±0.02} & \textbf{0.76±0.03} & \textbf{0.64±0.02} & \textbf{0.66±0.03} \\
\midrule
& Lift@K\% $e_x$
& \textit{2.91±0.24} & \textit{4.99±1.12}
& \textit{3.35±0.32} & \textit{5.20±0.94}
& \textit{3.46±0.35} & \textit{6.22±0.95}
& \textit{2.39±0.14} & \textit{3.75±0.59} \\

\bottomrule
\end{tabular}
}}
\end{table}

We evaluated how well our models predict the top 5\% and top 10\% learner-video interaction moments from video segment embeddings, using different embedding configurations (Table~\ref{tab:modalities}). All behaviors were predicted substantially above the random baseline of 0.5, with top-5\% moments generally being more predictable than top-10\% moments. ${\text{RewoundTo}}_v(t)$ achieved the highest performance (up to 0.76 AUC), closely followed by ${\text{PausedAt}}_v(t)$ (0.74) and ${\text{Watched}}_v(t)$ (0.7), whereas ${\text{SkippedFrom}}_v(t)$ was more challenging (0.66).

Models trained on the full embedding $e_x$ consistently performed among the best. The subset of Qwen-VL's 
$e_{TV,\{1,32,64\}}$ performed almost as well. Even when only using visual embeddings
$e_{V,\{1,32,64\}}$, performance degraded only slightly. In contrast, slide text $e_{S}$ contributed the least information, while transcript $e_{T}$ was already sufficient to moderately outperform the random baseline. The centerframe embedding $e_{F,t}$ slightly outperformed $e_{T}$; adding neighboring frames $e_{F,(t-10,t,t+10)}$ substantially improved performance. When using $e_x$, $\text{Lift@K\%}$ indicates that the top 5\% ${\text{RewoundTo}}_v(t)$ predictions outperform the random baseline by a factor of more than 6. Even the most challenging case, the top 10\% ${\text{SkippedFrom}}_v(t)$, shows a mean $\text{Lift@K\%}$ of 2.39.

\begin{figure}[t]
  \centering
  \includegraphics[width=\linewidth, trim=0.4cm 0.25cm 0.5cm 0.25cm, clip]{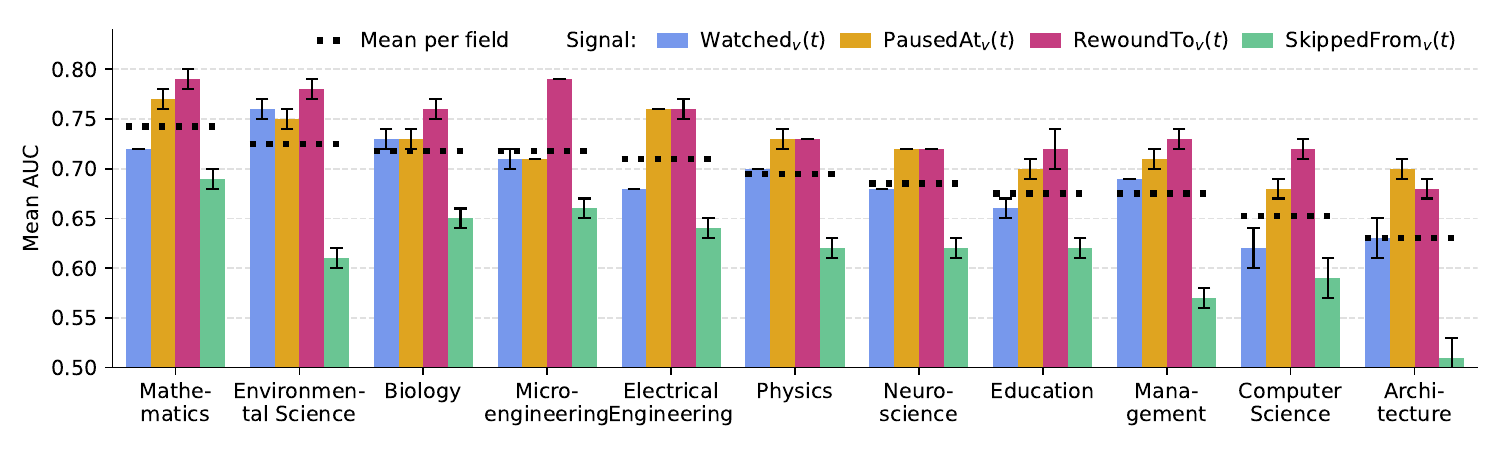}
  \caption{AUC ($\pm$ std, 5 seeds) for prediction of top 5\% learner-video interaction moments on fields unseen during training as a measure of generalization of our classifier.}
  \label{fig:fields_evaluation}
  \vspace{-3mm}
\end{figure}

Generalization to unseen academic fields is shown in Fig.~\ref{fig:fields_evaluation}. Across most STEM-adjacent fields, models generalized well with only modest performance drops due to domain shift. ${\text{PausedAt}}_v(t)$ and ${\text{RewoundTo}}_v(t)$ were more robust to domain shift compared to ${\text{Watched}}_v(t)$ and ${\text{SkippedFrom}}_v(t)$. Architecture and computer science were the most challenging held-out fields, whereas environmental science and management generalized comparatively well. Mathematics video interaction behavior was best predicted without training on such videos.

Table~\ref{tab:architecture} summarizes the role of the architecture, encoding depth, and model size. First, we tested not sharing weights in $h_1$ and removing the reference $\bar e_{x\in v}$. Including a reference increased the AUC by 2-4 percentage points, whereas sharing weights produced gains of 1-2 percentage points. When comparing different layers of Qwen-VL's 
$e_{TV}$ as input, layer 32 contained the most information across all signals. Interestingly, the 7B encoder version  
$e_{TV,\{1,32,64\}}$ performed almost as well as its larger 32B counterpart.

\begin{table}[t]
\centering
\caption{AUC ($\pm$ std, 10 seeds) for different neural architectures (top), three different layers of Qwen-VL (middle), and a smaller vision language model (bottom).}
\label{tab:architecture}
\resizebox{\textwidth}{!}{%
\rowcolors{3}{gray!7}{white}
\begin{tabular}{lllllllll}
\toprule
\multicolumn{1}{r}{\textbf{Signal}$\rightarrow$}
 & \multicolumn{2}{c}{\textbf{${\text{Watched}}_v(t)$}}
 & \multicolumn{2}{c}{\textbf{${\text{PausedAt}}_v(t)$}}
 & \multicolumn{2}{c}{\textbf{${\text{RewoundTo}}_v(t)$}}
 & \multicolumn{2}{c}{\textbf{${\text{SkippedFrom}}_v(t)$}} \\
\cmidrule(lr){2-3}
\cmidrule(lr){4-5}
\cmidrule(lr){6-7}
\cmidrule(lr){8-9}
\textbf{Setting}
& \textbf{Top 10\%} & \textbf{Top 5\%}
& \textbf{Top 10\%} & \textbf{Top 5\%}
& \textbf{Top 10\%} & \textbf{Top 5\%}
& \textbf{Top 10\%} & \textbf{Top 5\%} \\
\midrule
Reference $\bar e_x$ + Shared Weights in $h_1$ & \textbf{0.67±0.02} & \textbf{0.69±0.03} & \textbf{0.74±0.02} & \textbf{0.74±0.02} & \textbf{0.72±0.02} & 0.76±0.03 & \textbf{0.64±0.02} & \textbf{0.66±0.03} \\
Reference $\bar e_x$ + No sharing in $h_1$ & 0.66±0.02 & 0.68±0.03 & 0.72±0.02 & \textbf{0.74±0.02} & 0.71±0.02 & \textbf{0.77±0.03} & 0.61±0.03 & 0.65±0.03 \\
No Reference $\bar e_x$ & 0.66±0.02 & 0.68±0.03 & 0.70±0.03 & 0.71±0.03 & 0.71±0.02 & 0.73±0.03 & \textbf{0.64±0.03} & 0.63±0.04 \\
\midrule
1. Qwen-VL Layer $e_{TV,1}$ &  0.63±0.04 & 0.61±0.03 & 0.64±0.02 & 0.65±0.02 & 0.65±0.03 & 0.65±0.03 & 0.58±0.02 & 0.60±0.03 \\
32. Qwen-VL Layer $e_{TV,32}$ & \textbf{0.67±0.02} & \textbf{0.69±0.02} & 0.72±0.02 & 0.73±0.02 & \textbf{0.72±0.03} & 0.76±0.03 & 0.63±0.03 & 0.64±0.03 \\
64. Qwen-VL Layer $e_{TV,64}$ & 0.65±0.03 & 0.67±0.01 & 0.71±0.02 & 0.70±0.04 & 0.71±0.02 & 0.74±0.04 & 0.62±0.02 & 0.63±0.04 \\
\midrule
Small encoder $e_{TV,\{1,32,64\},7B}$ & \textbf{0.67±0.02} & \textbf{0.69±0.03} & 0.72±0.02 & 0.73±0.02 & 0.71±0.02 & 0.76±0.03 & 0.62±0.02 & \textbf{0.66±0.04} \\
\bottomrule
\end{tabular}
}
\vspace{-2mm}
\end{table}

\vspace{-2mm}
\subsection{RQ2: Predictiveness of AI-coded CTML Features}
\label{sec:rq2}
\vspace{-1mm}
We analyzed the informativeness of CTML features for the interaction signals. Due to the cost of coding these features, we focused on predicting the top 5\% of ${\text{PausedAt}}_v(t)$ and ${\text{RewoundTo}}_v(t)$, where our classifiers performed best.

\begin{figure}[t]
  \centering
  \includegraphics[width=\linewidth, trim=0.4cm 0.5cm 0.1cm 0.4cm, clip]{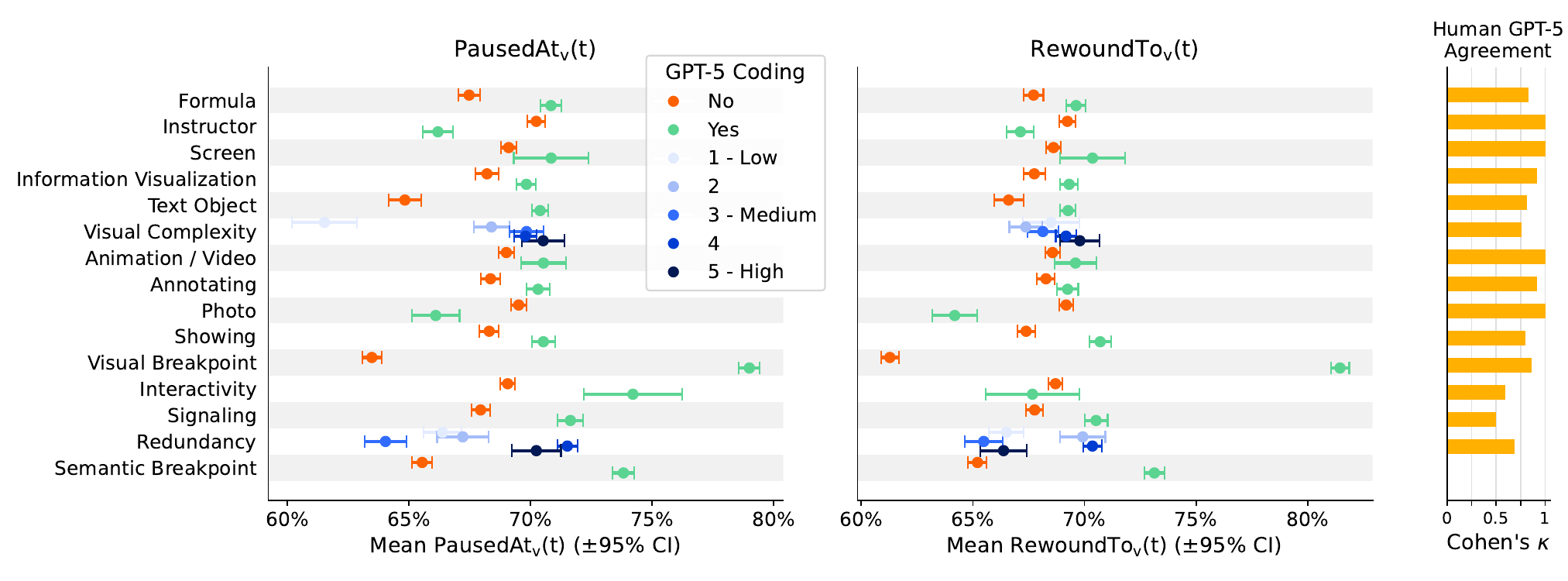}
  \caption{Learner-video interactions at 6,000 video moments (50\% from top 5\% ranks) by CTML features coded by GPT-5 (inter-rater agreement in the right panel). For example, moments without a formula have an average ${\text{PausedAt}}_v(t)$ rank of 67\% within their video, whereas moments with a formula have a higher average rank of 71\%.}
  \label{fig:rubric_effects}
\end{figure}

We first evaluated GPT-5’s ability to code CTML features at video moments. As shown in Fig.~\ref{fig:rubric_effects} (right), GPT-5 achieved almost perfect agreement  ($\kappa\ge0.8$) for 9 of 15 features and substantial agreement for three more features ($\kappa\ge0.6$). Agreement was weakest for the same categories where human annotators often disagreed, including the transcript-based features \textit{Interactivity} and \textit{Signaling} ($\kappa\ge0.4$), and at chance-level for \textit{Semantic Breakpoints} ($\kappa=0$). This may be due to the quality of the automatically created and translated transcripts, missing contexts, or their inherent subjectivity. We nonetheless retained all 15 features, as even low-agreement features may encode useful information.

Next, we examined associations between CTML features and interactions. As shown in Fig.~\ref{fig:rubric_effects}, most features were significantly related to the signals, with effects in both directions. For example, visual breakpoints were associated with more pauses and rewinds, whereas photos and instructor presence were associated with fewer of these events. Visual complexity exhibited a non-linear relationship with pausing, with substantially lower signals only at the lowest complexity level. We observed a similar pattern for redundancy, with the lowest signals co-occurring with medium redundancy. Despite low inter-rater agreement on semantic breakpoints, GPT-5 appeared to capture their presence at moments of higher pausing.

Finally, we analyzed the predictiveness of CTML features compared to multimodal embeddings. We used 5-fold cross-validation grouped by video due to the reduced sample size. Table~\ref{tab:ctml_prediction} compares the performance of our classifier using CTML features as the input vs. all embeddings $e_x$. In the small-sample regime ($<1,000$ datapoints), CTML features outperformed embeddings. Given $1,000-3,000$ datapoints, the two inputs appeared equivalent. The high-dimensional embeddings clearly outperformed CTML features at larger sample sizes; CTML feature performance plateaued at medium sample sizes.

\begin{table}[t]
\centering
\caption{Classifier performance (AUC$\pm$5-fold CV std) when using CTML features versus embeddings to predict learners’ interactions across training dataset sizes.}
\label{tab:ctml_prediction}
\resizebox{1\textwidth}{!}{%
\begin{tabular}{llllllllll}
\toprule
& \multicolumn{4}{c}{\textbf{CTML Features}} & \multicolumn{5}{c}{\textbf{Multimodal Embeddings} $e_x$} \\
\cmidrule(lr){2-5}\cmidrule(lr){6-10}
\textbf{Sample Size}$\rightarrow$ & \textbf{500} & \textbf{1,000} & \textbf{2,000} & \textbf{5,000} & \textbf{500} &\textbf{1,000} & \textbf{2,000} & \textbf{5,000} & \textbf{19,000} \\
\midrule

${\text{PausedAt}}_v(t)$ & 0.66 ± 0.02 & 0.69 ± 0.01 & 0.68 ± 0.01 & 0.68 ± 0.0 & 0.62 ± 0.02 & 0.67 ± 0.01 & 0.70 ± 0.0 & 0.72 ± 0.0 & \textbf{0.77 ± 0.01}\\
${\text{RewoundTo}}_v(t)$ & 0.73 ± 0.01 & 0.72 ± 0.0 & 0.73 ± 0.0 & 0.73 ± 0.0 & 0.68 ± 0.02 & 0.71 ± 0.01 & 0.73 ± 0.0 & 0.74 ± 0.0 & \textbf{0.78 ± 0.01}\\
\bottomrule
\end{tabular}
}
\vspace{-2mm}
\end{table}

\vspace{-2mm}
\subsection{RQ3: Sensitivity of Model Predictions to CTML features}
\vspace{-1mm}

\begin{figure}[t]
  \centering
  \includegraphics[width=1\linewidth,trim={0.125cm 2.25cm 0cm 2.5cm},clip]{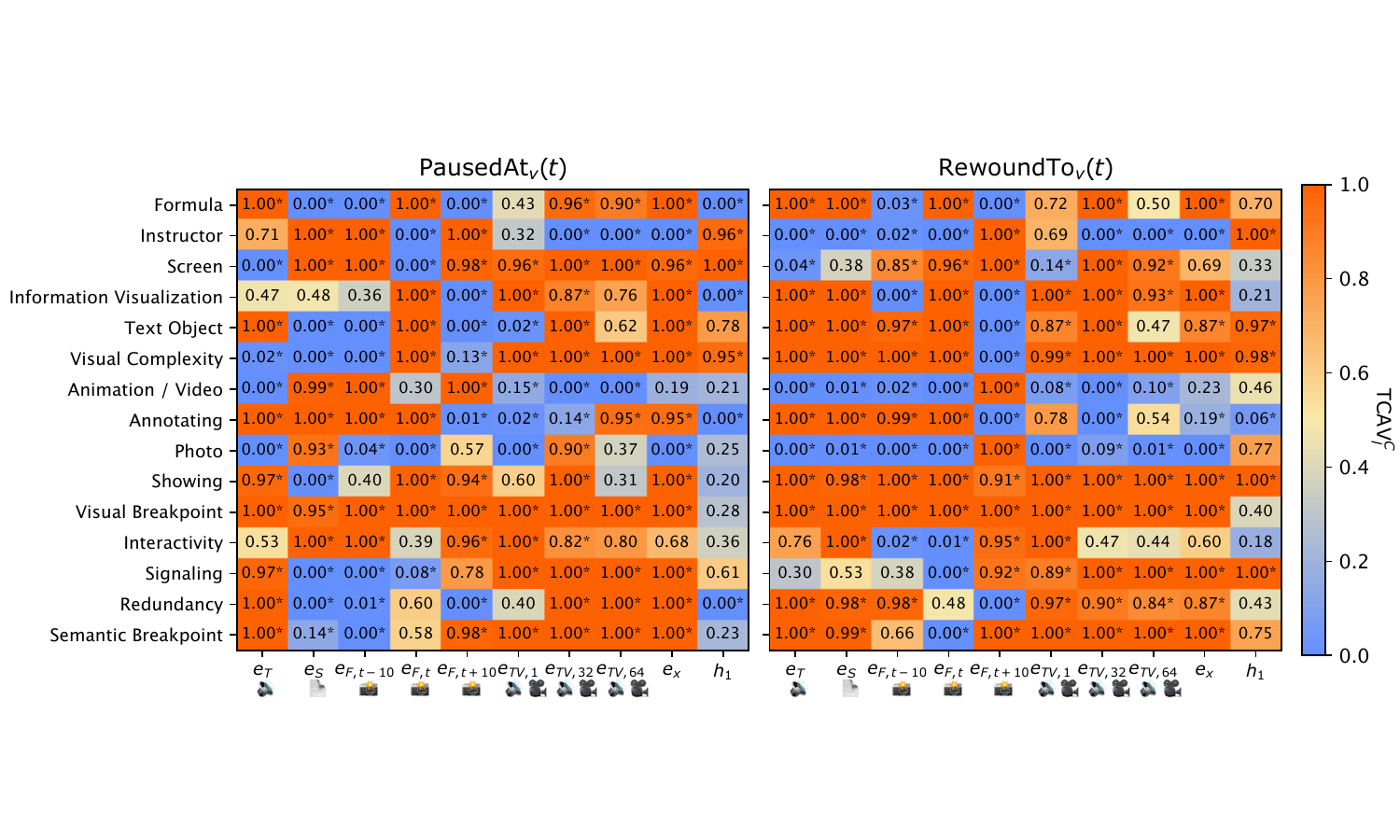}
  \caption{TCAV values of CTML concepts and activations in our classifier $H$. Significant (*) values above 0.5 mean that the classifier was positively sensitive to the concept present in the activation space.}
  \label{fig:tcav}
  \vspace{-4mm}
\end{figure}

Given the association between theory-driven CTML features and video interactions (see Fig.~\ref{fig:rubric_effects}), we tested whether $e_x$ encoded these features and whether the classifier $H$ was sensitive to these features. We computed TCAV values for individual embeddings, for their combination $e_x$, and for layer $h_1(e_x)$.

Fig.~\ref{fig:tcav} shows that the classifiers for both tested signals were sensitive to all CTML concepts. Across all concepts, at least one embedding space exhibited significant TCAV scores. For example, visual breakpoints yielded TCAV values close to 1 across all input embedding spaces. In contrast, this concept was less decisive in the hidden-layer representations, suggesting that it contributed mainly through the input embeddings rather than through intermediate abstractions. Moreover, for the ${\text{PausedAt}}_v(t)$ prediction, formulas and screens exhibited modality-dependent effects: formulas appearing in spoken content were positively associated with pausing, whereas representations in slide text showed a negative association. Finally, TCAV results on $e_x$ closely resembled the effects described in Fig.~\ref{fig:rubric_effects}. The high proportion of extreme TCAV scores (close to 0 or 1) may be explained by the binary classification compared to multiclass tasks.

When examining the linear separability of concepts in the input embeddings using cross-validation results from training the CAVs (see plot in \href{https://github.com/epfl-ml4ed/video-interaction-prediction-AIED26/blob/main/supplementary_material.pdf}{supplement}), we noted that especially the middle and last layer of Qwen2.5-VL linearly represented all concepts. Interestingly, even transcripts' embeddings contained information about visuals.
Considering the represented concepts, binary CTML concepts achieved an AUC of 0.82 in at least one modality, while classifiers for ordinal features showed a minimum 69\% improvement in MSE.

\vspace{-3mm}
\section{Discussion and Conclusion}
\vspace{-2mm}
Motivated by the lack of detailed behavioral feedback for improving video design before deployment, we developed and evaluated a scalable, explainable pipeline that links video content to learners’ temporally fine-grained interaction behavior. 

Addressing \textbf{RQ1}, we found that classifiers based on MLLM embeddings can reliably identify high-ranking interaction moments, achieving AUCs of up to 0.76. While watching, pausing, and rewinding behavior were predicted substantially above chance, predicting skipped moments was more challenging, indicating that forward seeking is less driven by instantaneous content and instead reflects longer-term viewing decisions beyond the 20-second context window. Encouragingly, the models generalized well to unseen academic fields within the STEM domain, likely due to overlapping instructional structures and shared content (e.g., common mathematical techniques across disciplines).

For \textbf{RQ2}, we found that GPT-5 achieved high agreement when coding visually grounded CTML features, matching human reliability where humans themselves agreed, but struggling with more subjective, transcript-based concepts such as signaling, interactivity, and semantic breakpoints. Notably, even these lower-agreement features showed significant differences in behavioral signals, suggesting that GPT-5 captures a consistent notion of these abstract concepts, which warrants further investigation. When comparing predictions from CTML features and embeddings, qualitative coding saturated quickly, whereas embeddings captured richer information at scale, despite incurring less than 1\% of the cost of OpenAI API usage when computed via a single forward pass of a medium-sized MLLM (Qwen2.5-VL).

For \textbf{RQ3}, TCAV revealed that the pausing and rewinding classifiers were sensitive to CTML features, particularly to visual concepts such as visual breakpoints, instructor presence, text objects, structured visualizations, annotations, and redundancy. These results demonstrate that multimodal embeddings capture interpretable, theory-relevant instructional concepts and that model predictions are meaningfully grounded in multimedia learning principles rather than opaque correlations. TCAV can therefore be used for explaining individual predictions.

\vspace{1pt} \noindent \textbf{Limitations}. Interaction signals do not directly indicate the quality of instructional design. However, our predictions can direct instructors' attention to potentially problematic video segments. In addition to the ${\text{RewoundTo}}_v(t)$ and ${\text{SkippedFrom}}_v(t)$ signals, the origin timestamp of rewinds and the target of skips should also be analyzed. Also, the behavioral prediction depends on the intended audience of a video, as our models were trained on a specific population. Finally, we only encoded CTML features to preserve generalizability, neglecting field-specific content features. Despite this generality, further evaluation on less related fields, such as the humanities, is required to fully assess generalization. 

\vspace{1pt} \noindent \textbf{Implications}. Our work has several implications for educational stakeholders. Universities and instructors launching new courses often lack sufficient historical learner data to guide design decisions. Our approach addresses this gap by predicting likely viewing behavior from video content only, removing the dependency on learner data. This is especially valuable for institutions with smaller enrollments, where large-scale data may never materialize. For instructors, the models provide early feedback on whether segments are watched as expected, along with theoretically grounded explanations.

\begin{credits}
\subsubsection{\ackname} This work was supported by the Jacobs Foundation as part of the consortium DEEP (Digital Education for Equity in Primary Schools). We thank the Center for Digital Education at EPFL for providing the data for this study. We thank Mrinmaya Sachan, Patrick Jermann, and Annechien Sarah Helsdingen for their helpful comments on the conceptualization and the manuscript.

\subsubsection{\discintname}
The authors have no competing interests to declare that are relevant to the content of this article.
\end{credits}
\bibliographystyle{splncs04}
\bibliography{references_short}

\end{document}